\documentclass[letterpaper]{article} 
\usepackage{aaai2026}  
\usepackage{times}  
\usepackage{helvet}  
\usepackage{courier}  
\usepackage[hyphens]{url}  
\usepackage{graphicx} 
\usepackage{natbib}  
\usepackage{caption} 
\usepackage{algorithm}
\usepackage{algorithmic}

\usepackage[table]{xcolor}
\usepackage{booktabs}   
\usepackage{graphicx}
\usepackage{subcaption}
\usepackage{amsmath}
\usepackage{adjustbox}
\usepackage{multirow}
\usepackage{makecell}
\usepackage{pifont}         
\usepackage{listings}
\lstdefinestyle{promptstyle}{
    backgroundcolor=\color{gray!10},       
    basicstyle=\ttfamily\small,             
    breaklines=true,                        
    breakindent=0pt,
    breakatwhitespace=true,                 
    columns=fullflexible,                   
    frame=single,                           
    captionpos=b,                           
    numbers=none,                           
    numberstyle=\tiny\color{gray},          
    keywordstyle=\color{blue!70},           
    commentstyle=\color{green!60!black},     
    stringstyle=\color{red!60!black},        
    showstringspaces=false                  
}

\usepackage{newfloat}
\usepackage{listings}
\DeclareCaptionStyle{ruled}{labelfont=normalfont,labelsep=colon,strut=off} 
\floatstyle{ruled}
\newfloat{listing}{tb}{lst}{}
\floatname{listing}{Listing}
\title{Connecting the Dots: Training‑Free Visual Grounding via Agentic Reasoning}
\author{
    Liqin Luo\textsuperscript{\rm 1}\equalcontrib,
    Guangyao Chen\textsuperscript{\rm 1}\equalcontrib\thanks{Corresponding authors.},
    Xiawu Zheng\textsuperscript{\rm 4},
    Yongxing Dai\textsuperscript{\rm 1},
    Yixiong Zou\textsuperscript{\rm 5},\\
    Yonghong Tian\textsuperscript{\rm 1,2,3}\footnotemark[2]
}
\affiliations{

    \textsuperscript{\rm 1}National Key Laboratory for Multimedia Information Processing, School of Computer Science, Peking University\\
    \textsuperscript{\rm 2}Peng Cheng Laboratory, China\\
    \textsuperscript{\rm 3}School of AI for Science, Peking University\\
    \textsuperscript{\rm 4}Institute of Artificial Intelligence, Xiamen University\\
    \textsuperscript{\rm 5}School of Computer Science and Technology, Huazhong University of Science and Technology\\
    
}

\usepackage{bibentry}

\begin{document}

\maketitle

\begin{links}
\end{links}

\begin{abstract}
Visual grounding, the task of linking textual queries to specific regions within images, plays a pivotal role in vision-language integration. Existing methods typically rely on extensive task-specific annotations and fine-tuning, limiting their ability to generalize effectively to novel or out-of-distribution scenarios. To address these limitations, we introduce \textbf{GroundingAgent}, a novel \textit{agentic visual grounding} framework that operates \textit{without any task-specific fine-tuning}. GroundingAgent employs a structured, iterative reasoning mechanism that integrates pretrained open-vocabulary object detectors, multimodal large language models (MLLMs), and large language models (LLMs) to progressively refine candidate regions through joint semantic and spatial analyses. Remarkably, GroundingAgent achieves an average zero-shot grounding accuracy of 65.1\% on widely-used benchmarks (RefCOCO, RefCOCO+, RefCOCOg), entirely without fine-tuning. Furthermore, by substituting MLLM-generated captions with the original query texts, the accuracy at the selection stage alone reaches approximately 90\%, closely matching supervised performance and underscoring the critical role of LLM reasoning capabilities. GroundingAgent also offers strong interpretability, transparently illustrating each reasoning step, thus providing clear insights into its decision-making process. The code is released on \url{https://github.com/loiqy/GroundingAgent}.
\end{abstract}

\section{Introduction}
\label{sec:intro}

Visual grounding (VG), the task of associating natural language descriptions with specific image regions, is crucial for bridging visual perception and language understanding. This capability underpins numerous downstream tasks, such as visual question answering, human-robot interaction, and interactive image retrieval. Traditional VG benchmarks, including Referring Expression Comprehension (REC)~\cite{refcoco, refcocog} and Referring Expression Segmentation (RES), typically rely on meticulously annotated datasets containing tens of thousands of image-object pairs. This limited scale contrasts sharply with contemporary object detection datasets, which commonly offer millions of annotated instances~\cite{lin2014microsoft,gupta2019lvis,zou2021annotation,zhang2024micm,xiao2025every}. Consequently, models trained on conventional VG datasets struggle to generalize to open-world scenarios, particularly in zero-shot conditions involving novel or out-of-distribution concepts. Addressing these challenges necessitates sophisticated semantic interpretation, comprehensive scene understanding, and precise spatial localization.

\begin{figure*}[t]
\centering
\includegraphics[width=\linewidth]{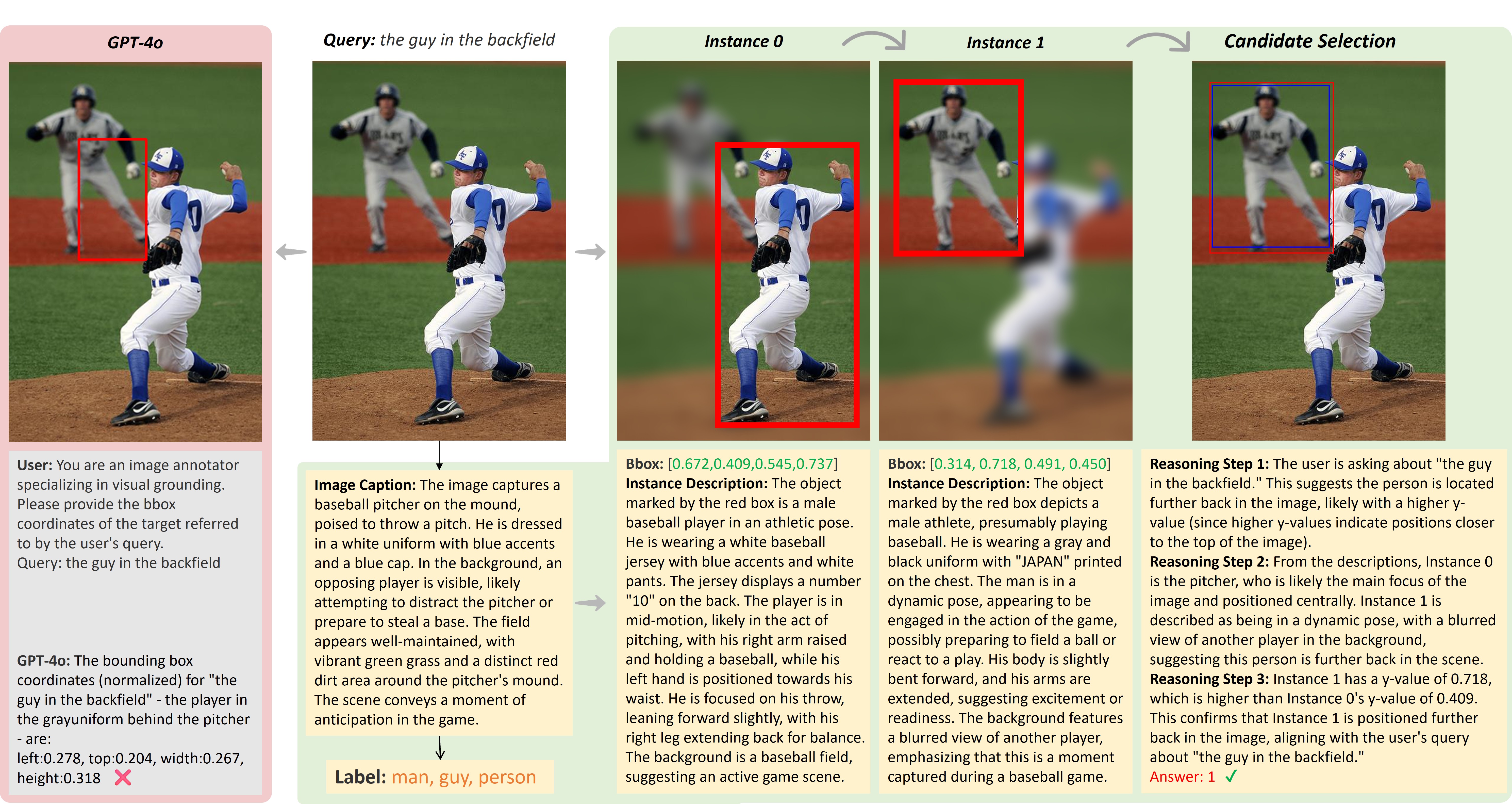}
\caption{Qualitative comparison and reasoning steps for the visual grounding task. Given the same image and query, the baseline GPT-4o prediction (red box) incorrectly selects the pitcher (left). Our method performs several iterative instance proposals to find the correct object through visual reasoning.} 
\label{fig:finding}
\end{figure*}

Recently, Transformer-based detectors such as Grounding DINO~\cite{liu2024grounding} have achieved strong results on visual grounding (VG) benchmarks, notably RefCOCO and RefCOCO+\cite{refcoco,refcocog}. In contrast, multimodal large language models (MLLMs)\cite{alayrac2022flamingo,zeng2025objects,chen2025automated}, while proficient in image captioning and visual question answering due to extensive image-text pre-training~\cite{zhai2022scaling}, exhibit poor localization performance without specialized VG training. For instance, GPT-4o struggles to accurately predict bounding boxes, as illustrated in Figure~\ref{fig:finding}, and language-centric models such as Kosmos-2~\cite{kosmos2} significantly underperform compared to detection-based models. Addressing this gap by acquiring additional VG-specific annotations is prohibitively expensive, as precise bounding-box or segmentation annotations are substantially more resource-intensive than image-level captions typically used for MLLM pre-training.

To overcome these challenges, we introduce \textbf{GroundingAgent}, a novel \textit{training-free} visual grounding framework empowered by \textit{agentic reasoning}. Unlike conventional methods that rely heavily on costly, task-specific fine-tuning, GroundingAgent capitalizes on the synergistic combination of pre-trained open-vocabulary object detectors, MLLMs, and LLMs. Specifically, GroundingAgent first utilizes an LLM to infer semantically relevant candidate concepts from the given textual query. Subsequently, these concepts guide an open-vocabulary object detector to generate candidate bounding boxes from the input image. Each candidate region is then enriched with detailed visual-semantic descriptions through joint multimodal analysis. These enriched candidates undergo a agentic reasoning process: an LLM progressively refines predictions by considering global image context, candidate semantics, and accumulated reasoning outputs.

We extensively evaluate GroundingAgent on widely-used benchmarks (RefCOCO, RefCOCO+, RefCOCOg), demonstrating superior performance in zero-shot grounding scenarios. GroundingAgent distinguishes itself through several notable advantages: \textbf{Firstly}, its open-vocabulary design naturally accommodates novel concepts without being constrained by predefined categories. \textbf{Secondly}, by leveraging pretrained MLLMs without any task-specific fine-tuning, GroundingAgent achieves an impressive average accuracy of 65.1\% in a fully training-free manner. \textbf{Thirdly}, its explicit agentic reasoning pipeline significantly enhances interpretability, transparently revealing each step of the grounding process. \textbf{Fourthly}, replacing MLLM-generated captions with query texts during the candidate selection stage boosts selection accuracy to approximately 90\%, approaching supervised performance levels. This underscores the critical role of the large language model's reasoning capability in visual grounding.

Our main contributions can be summarized as follows:
\begin{itemize}
    \item We propose \textbf{GroundingAgent}, the first fully training-free visual grounding framework leveraging a structured, agentic reasoning pipeline. It seamlessly integrates pretrained open-vocabulary detectors with multimodal and large language models, entirely avoiding task-specific fine-tuning.
    \item GroundingAgent achieves state-of-the-art zero-shot performance on standard benchmarks (RefCOCO, RefCOCO+, RefCOCOg), surpassing previous zero-shot methods by significant margins and setting a robust new baseline for training-free visual grounding.
    \item Our framework demonstrates remarkable interpretability and flexibility, effectively handling complex linguistic instructions involving detailed attributes, spatial relationships, and ambiguous references. Its modular design further allows effortless integration or upgrades of pretrained vision and language models.
\end{itemize}
\section{Related Work}
\label{sec:related}

\noindent
\textbf{Visual Grounding.}
Visual grounding links textual descriptions to image regions and is addressed in both supervised and zero-shot settings. Early methods fine-tuned CNN-based detectors in two-stage~\cite{hong_learning_2022} or one-stage frameworks~\cite{yu_mattnet_2018,yang2020resc}, achieving strong closed-set performance but limited open-set flexibility. The introduction of Vision Transformers (ViTs)~\cite{dosovitskiy_image_2021} enabled fully transformer-based models like TransVG~\cite{deng_transvg_2022}, which improve visual–language alignment. More recent work leverages vision–language pre-trained models such as MDETR~\cite{kamath_mdetr_2021} and Grounding-DINO~\cite{liu2024grounding} or multi-task architectures like UniTAB~\cite{yang_unitab_2022} and OFA~\cite{wang_ofa_2022} to boost open-set grounding, at the expense of greater data and compute requirements.

\noindent
\textbf{Multimodal Large Language Model.}
Multimodal large language models (MLLMs) unify vision and language for a range of tasks. Flamingo~\cite{alayrac_flamingo_2022} and BLIP-2~\cite{li_blip-2_2023} align visual and textual representations via cross-attention or Q-Former modules. These models also support instance-level understanding, enabling visual referring and grounding. For referring, GPT4RoI~\cite{zhang_gpt4roi_2024} and Position-Enhanced Visual Instruction Tuning~\cite{chen_position-enhanced_2023} map text prompts to specific regions. For grounding, Kosmos-2~\cite{peng_kosmos-2_2023} and Grounding-DINO~\cite{liu2024grounding} combine box-level detection with grounded pre-training. Explicit approaches inject location tokens~\cite{peng_kosmos-2_2023,wang_visionllm_2023}, while implicit methods leverage visual–textual cues~\cite{chen_shikra_2023}. Extending these techniques to coherent multi-round dialogues remains a key challenge~\cite{chen_shikra_2023}.

\section{Agentic Visual Grounding}
\label{sec:method}

\begin{figure*}[t]
    \centering
    \setlength{\abovecaptionskip}{0.1cm}
    \includegraphics[width=\linewidth]{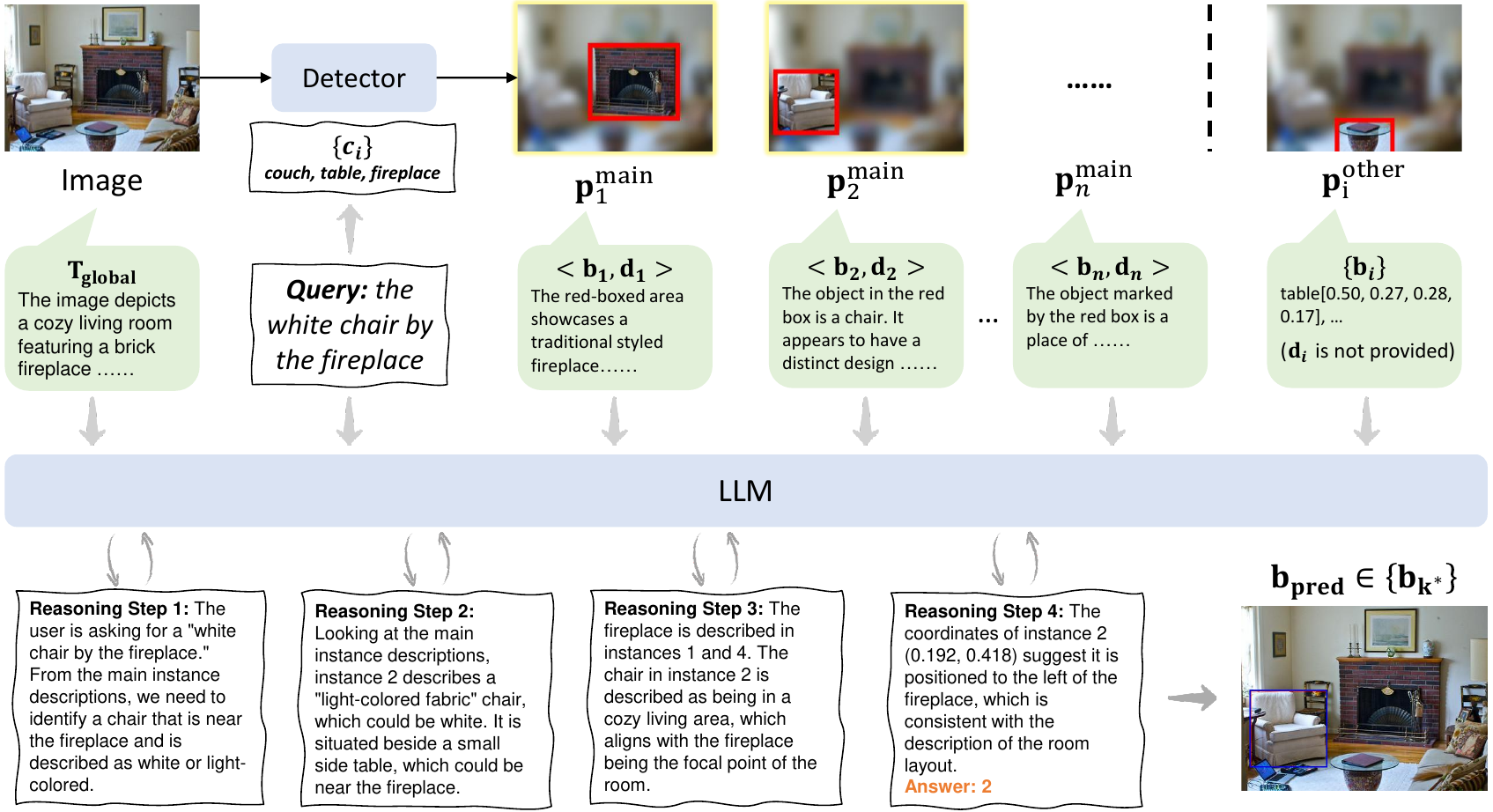}
    \caption{\textbf{Illustration of our step-by-step reasoning framework for zero-shot referring expression comprehension.} Given an input image and a textual query (e.g., \emph{``the white chair by the fireplace''}), the system first extracts a global description $\mathbf{T}_{\text{global}}$ of the scene and generates candidate bounding boxes ($\{\textbf{b}_i\}$) through an object detector. For each candidate region $\textbf{b}_i$, an MLLM is employed to generate a fine-grained semantic description $\textbf{d}_i$, capturing detailed visual attributes and contextual cues. These descriptions, along with the global context and the original query, are passed to an LLM, which performs step-by-step reasoning to refine its understanding of each candidate. In this example, four reasoning steps guide the LLM to identify and confirm the correct bounding box for the white chair, ensuring consistency with the spatial layout and visual attributes described in the query. The final prediction $\mathbf{b}_{\text{pred}}$ is chosen from the candidate set as the best match for the referring expression.
    }
    \label{fig:framework}
\end{figure*}

\paragraph{Problem Definition.}
Given an input image $I$ and a natural language query $Q$, visual grounding aims to locate the target object described by $Q$ by predicting its bounding box $\mathbf{b}_{\text{pred}}$. Formally, let $\mathcal{B}(I)$ denote all possible bounding boxes in $I$:
\[
  \mathbf{b}_{\text{pred}}
  =\arg\max_{\mathbf{b}\in\mathcal{B}(I)}\phi(I,Q,\mathbf{b}),
\]
where $\phi(I,Q,\mathbf{b})$ measures the alignment between the visual content in $\mathbf{b}$ and the semantic information in $Q$. Unlike conventional object detection, where the model simply recognizes and localizes pre-defined object categories, visual grounding requires a joint understanding of both visual cues and the linguistic nuances embedded in $Q$. Here, $f_{\text{vis}}(I,\mathbf{b})$ and $f_{\text{lang}}(Q)$ denote visual and linguistic representations, respectively, and we define
\[
  \phi(I,Q,\mathbf{b})
  =\operatorname{sim}\bigl(f_{\text{vis}}(I,\mathbf{b}),\,f_{\text{lang}}(Q)\bigr),
\]
with $\operatorname{sim}(\cdot,\cdot)$ as a similarity metric (e.g., cosine). The challenge is to optimize this objective in a zero-shot setting without task-specific fine-tuning.

\subsubsection{GroundingAgent}
We proposed, GroundingAgent, a training-free framework for zero-shot visual grounding.
As shown in Figure~\ref{fig:framework}, a pretrained open-vocabulary detector first suggests candidate bounding boxes.
A multimodal large language model (MLLM) then supplies rich semantic descriptions for each region.
Finally, a large language model (LLM) reasons step by step over these descriptions, spatial cues, and scene context to pick the box that best matches the textual query.
The whole pipeline works without task-specific fine-tuning and provides clear, interpretable reasoning traces.

\subsubsection{Candidate Generation}
For the process for generating candidate target regions, an image caption is firstly generated using an MLLM, which we denote as $C(I)$. This caption provides a comprehensive description of the image content. By concatenating the natural language query $Q$ with the generated caption $C(I)$, we form an enriched textual context that better reflects both the user’s intent and the semantic content of the image. Based on this enriched context, a large language model infers a set of semantically relevant candidate target concepts:
\begin{equation}
    \mathcal{C}(Q, C(I)) = \{ c_1, c_2, \ldots, c_{|\mathcal{C}(Q,C(I))|} \}.
\end{equation}

For each candidate concept $c \in \mathcal{C}(Q,C(I))$, we employ an open-vocabulary object detector on the input image $I$ to identify corresponding object instances. Specifically, for each concept $c$, the detector yields a set of candidate bounding boxes:
\begin{equation}
    \mathcal{D}_c(I) = \{ \mathbf{b}_{c,1}, \mathbf{b}_{c,2}, \ldots, \mathbf{b}_{c,m_c} \},
\end{equation}
where $\mathbf{b}_{c,j}$ denotes the $j$-th bounding box associated with concept $c$, and $m_c$ is the total number of detections for $c$. The union of all such detections across candidate concepts forms the overall candidate set:
\begin{equation}
    \mathcal{D}(I,Q,C(I)) = \bigcup_{c\in\mathcal{C}(Q,C(I))} \mathcal{D}_c(I).
\end{equation}

\begin{algorithm}[t]
    \small
    \caption{GroundingAgent: Training-Free Visual Grounding via Agentic Reasoning}
    \label{alg:groundingagent}
    \begin{algorithmic}[1]
        \REQUIRE Image $I$, natural language query $Q$
        \ENSURE Predicted bounding box $\mathbf{b}_{\text{pred}}$
        \medskip
        \STATE \textbf{Candidate Generation:}
        \STATE $C(I) \gets \text{MLLM}(I)$ \COMMENT{Generate global image description}
        \STATE $\mathbf{T}_{\text{global}} \gets C(I)$ \COMMENT{Obtain overall semantic context}
        \STATE $\tilde{Q} \gets \text{Concat}(Q,\, C(I))$ \COMMENT{Form enriched query context}
        \STATE $\mathcal{C} \gets \text{LLM}(\tilde{Q})$ \COMMENT{Infer candidate concepts from enriched query}
        \FOR{each concept $c \in \mathcal{C}$}
            \STATE $\mathcal{D}_c(I) \gets \text{Detector}(I,\, c)$ \COMMENT{Detect candidate bounding boxes for concept $c$}
        \ENDFOR
        \STATE $\mathcal{D}(I,Q,C(I)) \gets \bigcup_{c \in \mathcal{C}} \mathcal{D}_c(I)$
        \FOR{each candidate bounding box $\mathbf{b}_i \in \mathcal{D}(I,Q,C(I))$}
            \STATE $\mathbf{d}_i \gets f_{\text{desc}}(I,\, \mathbf{b}_i)$ \COMMENT{Generate detailed description}
        \ENDFOR
        \STATE $\begin{aligned}[t]
            &\mathcal{D}'(I,Q,C(I)) \gets \\
            &\quad \text{Sort}\Bigl(\mathrm{NMS}(\{(\mathbf{b}_i,\mathbf{d}_i)\}),\, \text{by } \text{area}(\mathbf{b}_i) \text{ (desc)}\Bigr)
        \end{aligned}$
        \STATE \textbf{Candidate Selection:}
        \FOR{each candidate $(\mathbf{b}_i, \mathbf{d}_i) \in \mathcal{D}'(I,Q,C(I))$}
            \STATE $r_i \gets \text{LLM}(Q,\, \mathbf{T}_{\text{global}},\, \mathbf{b}_i,\, \mathbf{d}_i)$ \COMMENT{Evaluate candidate }
        \ENDFOR
        \STATE $k^* \gets \text{find}\Bigl(\{r_i\},\, r_i=1\Bigr)$
        \RETURN $\mathbf{b}_{\text{pred}} \gets \mathbf{b}_{k^*}$
    \end{algorithmic}
\end{algorithm}

To further refine these candidates, we leverage an MLLM to jointly analyze the global image $I$ and each localized region defined by a bounding box $\mathbf{b}_i$. Let $f_{\text{desc}}(I, \mathbf{b}_i)$ denote the function that generates a detailed description $\mathbf{d}_i$ for the candidate region:
\begin{equation}
    \mathbf{d}_i = f_{\text{desc}}(I, \mathbf{b}_i),
\end{equation}
where $\mathbf{d}_i$ encapsulates both the visual characteristics and semantic attributes of the candidate.

Finally, to prioritize salient objects, the candidate bounding boxes are sorted in descending order of their areas. Denoting the area of a bounding box $\mathbf{b}_i$ as $\text{area}(\mathbf{b}_i)$, the sorted candidate set with non-maximum suppression (NMS) is given by:
\begin{equation}
    \label{eq:refine}
    \begin{aligned}
        \mathcal{D}'(I,Q,C(I)) = {} & \text{Sort}\Bigl(\mathrm{NMS}(\mathcal{D}(I,Q,C(I))),\, \\
        & \text{by } \text{area}(\mathbf{b}_i) \text{ (desc)}\Bigr).
    \end{aligned}
\end{equation}
After this refinement and sorting stage, each candidate is represented as a tuple $(\mathbf{b}_i, \mathbf{d}_i)$, which serves as the foundation for the subsequent grounding process.

\subsubsection{Candidate Selection}
After enriching and sorting the candidate set, we directly select the most appropriate candidate from $\mathcal{D}_{\text{ref}} = \mathcal{D}'(I,Q,C(I))$ based solely on its descriptive caption. For each candidate tuple $(\mathbf{b}_i, \mathbf{d}_i)$, the LLM is provided with the query $Q$, the global context $\mathbf{T}_{\text{global}}$, the candidate’s bounding box $\mathbf{b}_i$, and its detailed description $\mathbf{d}_i$. To further enhance decision-making reliability and interpretability, we incorporate a Chain-of-Thought (CoT)~\cite{wei2022chain,wang2023selfconsistencyimproveschainthought} reasoning module into the process. Specifically, the LLM generates intermediate reasoning steps that explicitly articulate the logical connections between the query, the global context, and the candidate’s description. This intermediate chain not only validates the semantic coherence of the candidate with respect to the query but also serves as an internal explanation for the final decision. Without computing any explicit confidence score, the LLM leverages both its semantic understanding and the CoT-derived reasoning to directly judge whether the candidate best corresponds to the query by evaluating its caption:
\begin{equation}
    \mathbf{r}_i = \text{LLM}\Bigl(Q,\; \mathbf{T}_{\text{global}},\; \mathbf{b}_i,\; \mathbf{d}_i\Bigr) \in \{0,1\},
\end{equation}
where $\mathbf{r}_i = 1$ indicates that the candidate is considered a match.
For traditional REC tasks like RefCOCO, which require referring to only one region, the output needs to be constrained to a one-hot format—meaning that there is a unique $\mathbf{r}_i = 1$, and all others are 0. Similarly, this framework can be easily extended to more general referring tasks, where the dataset may include tasks with no referred region or cases where a single referring query corresponds to multiple referred regions\cite{xia_gsva_2024, liu2023gres}.
In this way, our caption-based selection strategy not only capitalizes on the LLM's deep semantic understanding but also benefits from the transparent, step-by-step reasoning enabled by the CoT module, thereby obviating the need for iterative refinement or explicit confidence scoring.

\begin{table*}[!t]
    \centering
    \setlength{\abovecaptionskip}{0.1cm}
    \begin{adjustbox}{max width=\linewidth}
    \begin{tabular}{l|c|ccc|ccc|cc|c}
        \toprule
        \multirow{2}{*}{\textbf{Method}} & \multirow{2}{*}{\textbf{\makecell{Zero-shot}}} &  \multicolumn{3}{c|}{\textbf{RefCOCO}} & \multicolumn{3}{c|}{\textbf{RefCOCO+}} & \multicolumn{2}{c|}{\textbf{RefCOCOg}} & \multirow{2}{*}{\textbf{Avg}}\\
        & & \textbf{val} & \textbf{testA} & \textbf{testB} & \textbf{val} & \textbf{testA} & \textbf{testB} & \textbf{val} & \textbf{test} &\\
        \midrule
        Pseudo-Q~\cite{jiang2022pseudo} & \ding{55} & 56.0 & 58.3 & 54.1 & 38.9 & 45.1 & 32.1 &49.8 &47.4  &47.7\\
        Grounding-Dino~\cite{liu2024grounding} & \ding{55} & 50.4 & 57.2 & 43.2 & 51.4 & 57.6 &45.8 & 67.5 & 67.1 & 55.0 \\
        MM-G~\cite{zhao2024open} & \ding{55} & 53.1 & 59.1 & 46.8 & 52.7 & 58.7 & 48.4 &62.9 &62.9 &55.6 \\
        Kosmos-2~\cite{kosmos2} & \ding{55} & 52.3 &57.4 &47.3 &45.5 &50.7 &42.2 &60.6 &61.7 &52.2 \\
        CoVLM~\cite{li2024covlm} & \ding{55} & 48.2 &53.2	&43.2	&47.6	&50.9	&44.2	&60.9	&61.9	&51.3 \\
        GLIP~\cite{li2021grounded} & \ding{55} & 50.4 & 54.3 & 43.8 & 49.6 & 52.8 &44.6 &66.1 &66.9 &53.6 \\
        REG~\cite{wang2024learning} & \ding{55} & \underline{63.4} & 68.5 & \underline{57.6} & 53.9 & 60.9 &44.9 &63.3 &63.2 & 59.5 \\ 
        \midrule
        CPT~\cite{yao2021cpt}  & \ding{51} & 32.2 & 36.1 & 30.3 & 31.9 & 35.2 & 28.8 &36.7 &36.5  &33.5\\
        VGDiffZero~\cite{liu2023vgdiffzero} & \ding{51} & 28.0 & 30.3 & 29.1 & 28.4 & 30.8 & 29.8 &33.5  &33.2 &33.9\\
        ReCLIP~\cite{subramanian2022reclip} & \ding{51} & 45.8 & 46.7 & 45.2 & 45.3 & 48.5 & 42.7 &57.0 &56.2  &48.4\\
        Red Circle$^*$~\cite{shtedritski2023does}  & \ding{51} & 49.8 & 58.6 & 39.9 & 55.3 & 63.9 & 45.4 & 59.4 & 58.9 & 53.9\\
        RelVLA~\cite{han2023zero}   & \ding{51} & 52.5 & 52.7 & 52.9 & 50.8 & 53.4 & 47.6 &61.3 &60.9  &54.0\\
        FGVP~\cite{yang2024fine}    & \ding{51} & 59.6 & 65.0 & 52.0 & 60.0 & 66.8 & 49.7 & 63.3 & 63.4 & 60.0 \\
        GroundVLP~\cite{shen2023groundvlpharnessingzeroshotvisual}  & \ding{51} & 59.1 & \underline{69.2} & 48.7 & \underline{61.8} & \textbf{70.6} & \underline{51.0} & \textbf{69.1} & \textbf{69.0} & \underline{62.3} \\
        \rowcolor{cyan!10}
        GroundingAgent & \ding{51} & \textbf{67.1} & \textbf{73.3} & \textbf{60.1} & \textbf{62.4} & \underline{67.6} & \textbf{53.8} & \underline{67.9} & \underline{68.8} & \textbf{65.1} \\ 
        \bottomrule
    \end{tabular}
    \end{adjustbox}
    \caption{Comparison with state-of-the-art methods on zero-shot referring expression comprehension (REC) tasks on RefCOCO/+/g dataset. "Zero-shot" here is defined as methods that do not use any task-specific grounding annotations (including manually labeled image-text corresponding regions, grounding labels, etc.) for training or fine-tuning, relying solely on pre-trained model capabilities for inference. The best two results are \textbf{bold-faced} and \underline{underlined}, respectively. }
    \label{tab:sota_rec}
\end{table*}

The overall process is summarized in Algorithm~\ref{alg:groundingagent}. Our proposed \text{GroundingAgent} leverages the complementary strengths of LLMs and open-vocabulary object detectors to perform zero-shot visual grounding. Specifically, GroundingAgent first generates candidate target concepts from the query, then enriches each candidate with detailed semantic and visual descriptions. A two-stage agentic reasoning module subsequently evaluates the spatial and semantic relationships among these candidates, enabling the robust identification of the target object described in $Q$, all without any additional training.

\section{Experiments}
\label{sec:exp}

\paragraph{Experiment Setting.}
To thoroughly evaluate the effectiveness of our proposed agent framework under zero-shot conditions, we conduct experiments on three widely adopted visual grounding benchmarks: RefCOCO~\cite{refcoco}, RefCOCO+~\cite{refcoco}, and RefCOCOg~\cite{refcocog}. These datasets provide a diverse set of referential expressions and corresponding images, facilitating a comprehensive assessment of both bounding-box-level and segmentation-level grounding capabilities. 
Specifically, we consider the standard task: Referring Expression Comprehension (REC). We measure the top-1 accuracy, where a prediction is deemed correct if the Intersection-over-Union (IoU) between the predicted bounding box and the ground-truth bounding box exceeds 0.5. 

\paragraph{Implementation Details.}
Our training-free visual grounding framework integrates four open-vocabulary detectors—APE~\cite{shen2024aligning}, Grounding DINO~\cite{liu2024grounding}, OWL-ViT~\cite{minderer2023scaling}, and YOLO World~\cite{Cheng2024YOLOWorld}—without any task-specific fine-tuning. By default, we use YOLO World, which is not trained on RefCOCO, ensuring an unbiased zero-shot evaluation. Candidate bounding boxes generated by YOLO are sorted by area to prioritize prominent regions.
We use Llama-3.2-11B-Vision~\cite{dubey_llama_2024} for generating both global and region-level visual descriptions. Semantic reasoning is performed by DeepSeek-V3 (\texttt{0324})~\cite{deepseekai2025deepseekv3technicalreport}, which iteratively extracts and refines relevant concepts from queries, global descriptions, and regional semantics. To enhance robustness, candidate regions smaller than 2.5\% of the image area are filtered out, and we retain a maximum of 10 primary candidates per image following non-maximum suppression. Additional details, including prompts, normalization strategies, and hyperparameters, are available in the Appendix.

\subsection{Main Results}
We present a comprehensive evaluation of our proposed \textit{GroundingAgent} framework, comparing its performance against state-of-the-art methods for zero-shot referring expression comprehension (REC) tasks on the widely-used RefCOCO, RefCOCO+, and RefCOCOg benchmarks. The detailed comparison results are summarized in Table~\ref{tab:sota_rec}.
GroundingAgent consistently achieves superior performance across all benchmark subsets, demonstrating clear advantages over recent zero-shot methods that do not utilize grounding annotations during training. 
Specifically, our framework attains an average accuracy of \textbf{65.1\%}, significantly outperforming fully zero-shot competitors such as VGDiffZero~\cite{liu2023vgdiffzero}, ReCLIP~\cite{subramanian2022reclip}, and Red Circle~\cite{shtedritski2023does}, with accuracy improvements of approximately 12-27\% across different subsets.
We further emphasize that while REG~\cite{wang2024learning} is not a zero-shot method, as it involves training with synthetically generated grounding annotations, introducing implicit supervision and thus deviating from a strictly training-free approach.
Despite this implicit training advantage, GroundingAgent outperforms REG on all evaluation splits, especially on challenging subsets like RefCOCO+ testB (53.8\%) and RefCOCOg test (68.8\%), underscoring the genuine zero-shot nature and effectiveness of our fully inference-based method.
Moreover, GroundingAgent achieves the best overall performance in terms of average accuracy (65.1\%), highlighting its robust generalization capabilities. This consistent performance improvement illustrates the effectiveness of integrating explicit semantic reasoning and structured agentic decision-making, enabling GroundingAgent to effectively interpret complex linguistic queries involving detailed attributes, ambiguous references, and sophisticated spatial relationships. Overall, these results clearly indicate GroundingAgent’s effectiveness and its potential as a powerful, flexible baseline for future advancements in the zero-shot visual grounding task.

\begin{figure}[t]
    \centering
    \setlength{\abovecaptionskip}{0.0cm}
    \includegraphics[width=\linewidth]{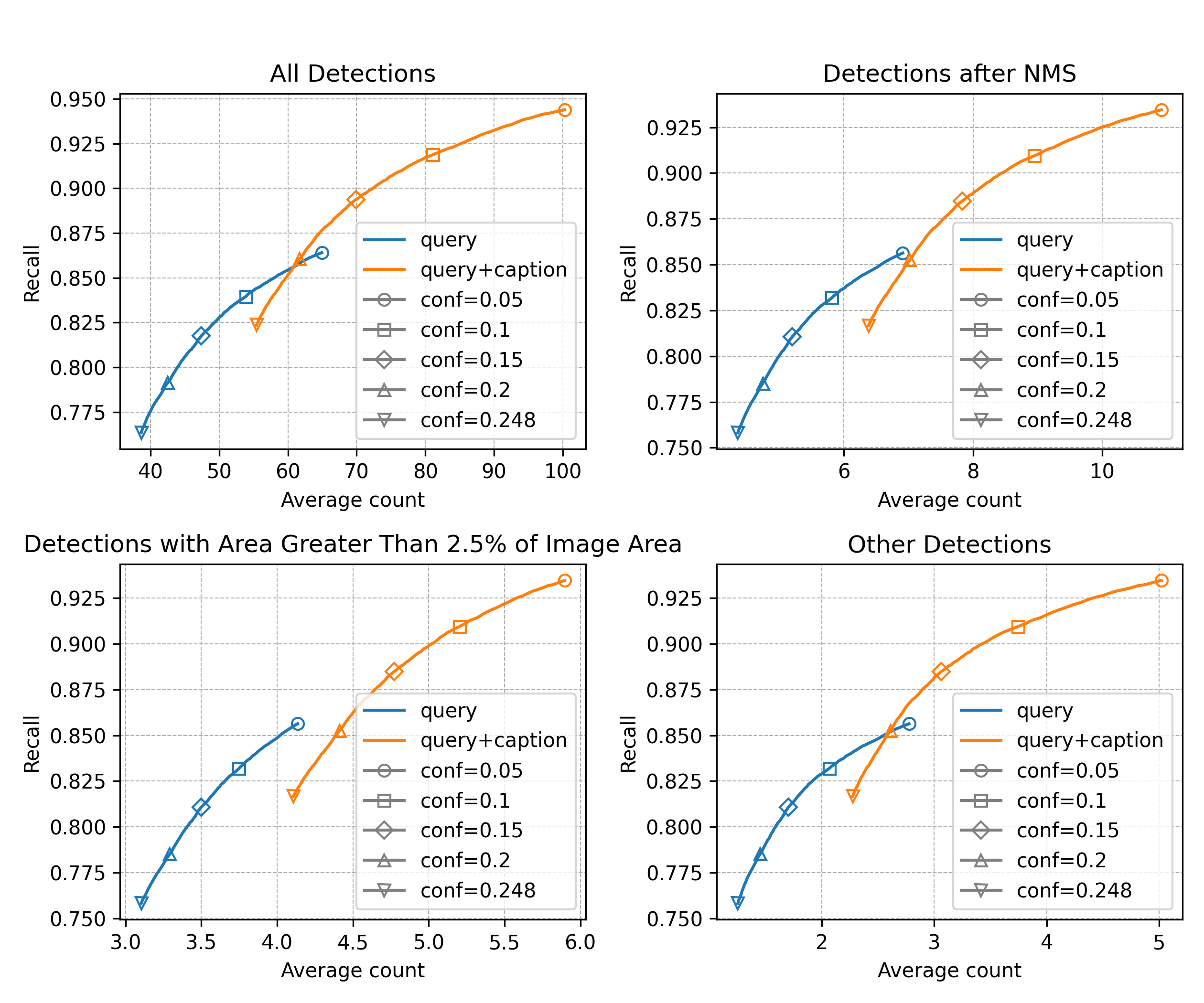}
    \caption{The recall of candidate generation on RefCOCO.} 
    \label{fig:hit}
\end{figure}

\begin{table*}[tbp]
    \centering
    \setlength{\abovecaptionskip}{0.1cm}
    \renewcommand{\arraystretch}{1.1} 
    \begin{adjustbox}{max width=\linewidth}
    
    \begin{tabular}{m{2.15cm}l|ccc|ccc|cc|c}
        \toprule
        \multirow{2}{*}{\textbf{Stage}} & \multirow{2}{*}{\textbf{Method}}
        & \multicolumn{3}{c|}{\textbf{RefCOCO}}
        & \multicolumn{3}{c|}{\textbf{RefCOCO+}}
        & \multicolumn{2}{c|}{\textbf{RefCOCOg}} & \multirow{2}{*}{\textbf{Avg}}\\
        & & \textbf{val} & \textbf{testA} & \textbf{testB}
          & \textbf{val} & \textbf{testA} & \textbf{testB}
          & \textbf{val} & \textbf{test} & \\
        \midrule
        \multirow{4}{*}{\makecell{Candidate\\Generation}}
        & APE~\cite{shen2024aligning}           & 98.6 & 98.7 & 97.9 & 98.1 & 98.5 & 98.0 & 98.4 & 98.6 & 98.3\\
        & GroundingDINO~\cite{liu2024grounding} & 98.3 & 98.7 & 97.6 & 97.9 & 99.0 & 97.8 & 98.2 & 98.3 & 98.2\\
        & OWL~\cite{minderer2023scaling}        & 95.7 & 96.3 & 92.6 & 95.7 & 95.7 & 93.0 & 95.0 & 95.3 & 94.9\\
        & YOLO-World~\cite{Cheng2024YOLOWorld}  & 94.4 & 96.7 & 91.1 & 94.2 & 95.7 & 91.3 & 93.2 & 93.8 & 93.8\\
        \midrule
        \multirow{3}{*}{\makecell{Candidate\\Selection}}
        & Caption
                & 67.1 & 73.3 & 60.1
                & 62.4 & 67.6 & 53.8
                & 67.9 & 68.8 & 65.1\\
        & \cellcolor{cyan!10}Caption $\rightarrow$ Query+
                & \cellcolor{cyan!10}82.9 & \cellcolor{cyan!10}83.7 & \cellcolor{cyan!10}78.8
                & \cellcolor{cyan!10}85.9 & \cellcolor{cyan!10}82.5 & \cellcolor{cyan!10}86.4
                & \cellcolor{cyan!10}89.6 & \cellcolor{cyan!10}90.2 & \cellcolor{cyan!10}85.0\\
        & \cellcolor{orange!10}Caption $\rightarrow$ Query
                & \cellcolor{orange!10}91.4 & \cellcolor{orange!10}91.4 & \cellcolor{orange!10}86.6
                & \cellcolor{orange!10}91.4 & \cellcolor{orange!10}89.9 & \cellcolor{orange!10}90.2
                & \cellcolor{orange!10}91.8 & \cellcolor{orange!10}92.4 & \cellcolor{orange!10}90.6\\
        \midrule
        \multirow{2}{*}{\makecell{Supervised\\SOTA}}
        & \cellcolor{cyan!10}Fine-tuned Model~\cite{Zheng_Zhao_Zheng_He_Cheng_Cai_Huang_2025}
                & \cellcolor{cyan!10}90.5 & \cellcolor{cyan!10}91.7 & \cellcolor{cyan!10}88.0 & \cellcolor{cyan!10}80.1 & \cellcolor{cyan!10}84.6 & \cellcolor{cyan!10}72.6 & \cellcolor{cyan!10}82.5 & \cellcolor{cyan!10}82.9 & \cellcolor{cyan!10}84.1\\
        & \cellcolor{orange!10}Pre-trained Model~\cite{bai2025qwen25vltechnicalreport}
                & \cellcolor{orange!10}92.7 & \cellcolor{orange!10}94.6 & \cellcolor{orange!10}89.7 & \cellcolor{orange!10}88.9 & \cellcolor{orange!10}92.2 & \cellcolor{orange!10}83.7 & \cellcolor{orange!10}89.9 & \cellcolor{orange!10}90.3 & \cellcolor{orange!10}90.3\\
        \bottomrule
    \end{tabular}
    
    \end{adjustbox}
    \caption{Ablation study on candidate generation and selection strategies. We report accuracy (\%) for each open-vocabulary detector independently, as well as the results after applying the candidate selection stage.}
    \label{tab:ablation}
\end{table*}

\subsection{Candidate Generation}
We conduct a detailed analysis to validate the effectiveness of our candidate generation stage, primarily using recall as our evaluation metric. Specifically, a candidate set is considered successful if the ground-truth bounding box is present among the generated candidate boxes. 

\paragraph{Robust candidate generation across detectors.}
First, we investigate the performance differences when employing various open-vocabulary detection models, including APE, YOLO-World, OWL, and GroundingDINO. The results, presented in Table~\ref{tab:ablation}, indicate that almost all detection models achieve recall rates around 95\%, underscoring the robustness of our candidate generation process. It is important to highlight that these results are obtained solely from bounding box outputs generated by detection models. We do not leverage any additional scoring, confidence metrics, or supplementary information from these detection models. Additionally, the inputs provided to these detection models are limited strictly to candidate vocabularies extracted via our proposed method, demonstrating the strength of the vocabulary generation step.

\begin{table}[tbp]
    \centering
    \setlength{\abovecaptionskip}{0.1cm}
    \begin{adjustbox}{max width=\linewidth}
    \begin{tabular}{l|cc}
        \toprule
         \multirow{2}{*}{\textbf{LLM}}
         & \multicolumn{2}{c}{\textbf{RefCOCO}} \\
        & \textbf{testA} & \textbf{testB} \\
        \midrule
        DeepSeek-V3~\cite{deepseekai2025deepseekv3technicalreport} & 73.3 & 60.1 \\
        DeepSeek-R1~\cite{deepseekai2025deepseekv3technicalreport} & 75.9 & 60.3 \\
        Llama3.1-8B~\cite{dubey_llama_2024} & 55.0 & 44.0 \\
        DeepSeek-R1-Llama-8B~\cite{deepseekai2025deepseekv3technicalreport} & 59.7 & 47.7 \\
        Qwen2.5-7B~\cite{qwen2025qwen25technicalreport} & 52.0 & 41.6 \\
        \bottomrule
    \end{tabular}
    \end{adjustbox}
    \caption{Ablation study on LLM in candidate selection stage.}
    \label{tab:ablation_llm}
\end{table}

\paragraph{Global captions are essential for precise vocabularies.}
To further examine the influence of different textual inputs during candidate vocabulary generation, we conduct experiments using two distinct input scenarios: using only the textual query versus combining both the textual query and a global image caption. The comparative results are illustrated in Figure~\ref{fig:hit}. It is evident from the figure that removing caption information leads to a significant drop in recall, highlighting the importance of global contextual captions. This suggests that incorporating global image descriptions effectively constrains the vocabulary generation process within the LLM, mitigating semantic divergence. Consequently, the captions guide the LLM toward generating more precise and relevant candidate vocabularies, directly enhancing the downstream grounding performance.

\subsection{Candidate Selection}

\paragraph{Caption quality is the main bottleneck.}
Table~\ref{tab:ablation} demonstrates a noticeable decline in overall performance following the candidate selection step. To thoroughly investigate this phenomenon, we conducted an ablation experiment where the caption descriptions generated by the MLLM for target instances were directly replaced with their original textual queries. Table~\ref{tab:ablation} also provides a detailed comparison illustrating that when using either the original query or a closely related textual description (Query+), the LLM achieves approximately 90\% accuracy. This result suggests that the candidate selection performance drop does not originate from our LLM-based selection process. Instead, it is largely due to inaccuracies in the caption generated by the MLLM.

\paragraph{Performance Alignment with Supervised SOTA.}
The performance of our framework after substituting MLLM-generated captions with Query+ (85.0\% average accuracy) closely approaches the accuracy of SOTA fine-tuned model~\cite{Zheng_Zhao_Zheng_He_Cheng_Cai_Huang_2025} (84.1\% average accuracy). This alignment suggests that when provided with query-enhanced contextual cues, our training-free framework can achieve results comparable to methods specifically optimized for visual grounding through task-specific fine-tuning, implicitly validating its ability to leverage semantic alignment without explicit supervision.
Furthermore, replacing captions with the original query directly yields an average accuracy of 90.6\%, which matches the performance of SOTA pre-trained model~\cite{bai2025qwen25vltechnicalreport} (90.3\% average accuracy) that are pretrained on massive vision-language datasets. This equivalence demonstrates that our framework’s core reasoning mechanism—when unimpaired by MLLM caption noise—can fully harness the semantic understanding capabilities of pre-trained models, confirming the effectiveness of its modular design in bridging vision and language without task-specific training.

\paragraph{Reasoning ability matters more than model size.}
We further explored the impact of different LLMs on selection performance through additional ablation studies (Table~\ref{tab:ablation_llm}). The results reveal significant performance disparities across LLMs: advanced models like DeepSeek-R1 achieve 75.9\% and 60.3\% accuracy on RefCOCO testA and testB, respectively, outperforming DeepSeek-V3. Notably, even among models with comparable parameter scales, performance varies substantially based on reasoning capabilities. For example, DeepSeek-R1-Llama-8B—which incorporates explicit reasoning training—outperforms the base Llama3.1-8B by 4.7-3.7 percentage points. This indicates that enhancing reasoning abilities through targeted training can effectively boost performance even without increasing model size, highlighting the critical role of structured reasoning in the selection process. In contrast, LLMs with weaker understanding and reasoning capabilities, such as Qwen2.5-7B, exhibit substantially lower performance.

\paragraph{Robust LLMs fix issues.}
For the reasoning steps, we conducted a qualitative analysis on the RefCOCO dataset. For LLMs with weak long-text processing, poor instruction-following, and limited reasoning performance, two issues may arise: first, \textbf{not reasoning}, meaning the LLM directly provides an answer without the necessary reasoning steps; second, \textbf{ambiguity}, meaning the descriptions generated by the LLM are too vague or unclear, sometimes even exhibiting inherent hallucinations in long text processing, which makes subsequent selection steps difficult. For high-performing LLMs, the model provides reasonable reasoning steps and clear descriptive information, effectively guiding the subsequent selection process. The average number of reasoning steps is 3.4, indicating that the LLM considers multiple steps during reasoning. We also found that, during the reasoning process, the LLM integrates the global semantic description of the image to infer or correct unreasonable text queries, thereby enhancing interpretability and improving robustness.

\subsection{Further Analysis}
\paragraph{Failure Analysis.}
Our method demonstrates robust performance, with quantitative analysis revealing low rejection rates on the RefCOCO+ benchmark (0.77\% on val, 0.73\% on testA, and 1.69\% on testB), confirming the effectiveness of our rejection-aware selection strategy. Primary error sources include inaccurate descriptions from multimodal generation artifacts and incorrect rejections with ambiguous referring expressions. Detailed failure case analysis is provided in the appendix.

\paragraph{Better explainability.}
To demonstrate interpretability, we present examples of accepted and rejected bounding boxes in Figure~\ref{fig:finding}. First, our visual prompt guides attention by outlining each candidate region in red and blurring the background\footnote{Gaussian blur with a standard deviation of 10.0 was applied.}. Then, our LLM-based selection strategy provides clear reasoning for each decision, explaining why candidates are accepted or rejected. When candidates share visual or spatial similarities, the framework evaluates their descriptions and positions to justify the final choice. This clarity enhances transparency, simplifies debugging, and reveals the model’s decision process.

\section{Conclusion}
\label{sec:con}
We presented \textit{GroundingAgent}, a training-free visual grounding framework that integrates pretrained models without task-specific fine-tuning. It achieves competitive zero-shot results on standard benchmarks and effectively interprets complex linguistic queries involving attributes, spatial relations, and ambiguity. The explicit reasoning pipeline enhances interpretability, while the modular design enables easy model substitution and future scalability. Our analysis shows that remaining limitations mainly arise from the fine-grained visual abilities of current MLLMs, suggesting clear directions for progress as multimodal modeling improves. Similar training-free, reasoning-based frameworks may also generalize well to other domains~\cite{yuan2025empowering,chen2025future} that require transparent and reliable decision-making.

\section*{Acknowledgements}
The authors gratefully acknowledge support from the Shenzhen High-Level Talent Team Program and the Shenzhen Science and Technology Innovation Commission (Grant No. KQTD 20240729102051063); the National Natural Science Foundation of China (Grant Nos. 62332002, 62027804, 61825101, and 62402015); the China Postdoctoral Science Foundation (Grant No. 2024M750100); and its Postdoctoral Fellowship Program (Grant No. GZB20230024). Computational resources were provided by Pengcheng Cloudbrain.

\bibliography{aaai2026}

\newpage
\newpage
\appendix

\section{Candidate Generation with Image Caption}

In Figure 3, we systematically analyze our candidate generation strategy by comparing bounding-box recall under different candidate selection conditions. We evaluate three distinct candidate subsets: (1) All detections before applying non-maximum suppression (NMS), (2) Detections after applying NMS, and (3) Major bounding boxes defined as detections occupying at least 2.5\% of the total image area. For all detections before NMS, integrating global image captions with textual queries consistently improves recall across varying confidence thresholds, achieving approximately 95\% recall at higher bounding-box counts. After applying NMS, recall remains robust despite significantly reduced bounding-box counts, again showing clear improvements when combining query and caption information. Moreover, we specifically analyze bounding boxes with larger visual prominence (area greater than 2.5\% of the image area). This subset achieves notably high recall performance, even with fewer candidates, further underscoring the effectiveness of caption-guided detection. For smaller detections, we observe their quantity increases as confidence thresholds decrease, which serves as an important indicator for threshold selection - we intentionally suppress these detections through proper thresholding to maintain focus on visually prominent objects. Overall, these results highlight the clear benefits of incorporating caption information in candidate generation, significantly enhancing recall performance while demonstrating our strategy's effectiveness in prioritizing relevant detections through adaptive threshold control.

\section{Effect of Small Ground-Truth Regions}

In the main text we reported recall across several open-vocabulary detectors. Here we provide a focused analysis that further explores how these four OVD variants behave specifically on small ground-truth targets. The goal is to investigate performance differences between detectors when GT instance area falls below different relative thresholds, rather than to justify our filtering choice.

\begin{table}[h]
    \centering
    \begin{adjustbox}{max width=\linewidth}
    \begin{tabular}{l|c|c}
        \toprule
        \textbf{OVD} & \textbf{$\beta = 5\%$} & \textbf{$\beta = 20\%$} \\
        \midrule
        APE   & 96.9 & 98.1 \\
        GD    & 95.3 & 96.9 \\
        OWL   & 94.9 & 93.9 \\
        YOLO  & 90.2 & 92.5 \\
        \bottomrule
    \end{tabular}
    \end{adjustbox}
    \caption{Performance of four OVD variants on small ground-truth instances (area $< \beta\%$ of image). The table highlights relative robustness to small-target detection across detectors.}
    \label{tab:ovd_small_targets}
\end{table}

The results indicate that while absolute differences between the four OVD variants are modest on aggregate, their relative behavior diverges when evaluating very small ground-truth instances. In particular, most GT boxes in our referring-expression datasets exceed the 2.5\% image-area scale used in the main pipeline, and the detectors differ in robustness to sub-5\% targets: some backbones retain high performance even on tiny instances, whereas others show larger drops. This focused analysis emphasizes that sensitivity to very small referents is a general challenge across detectors and motivates future work on proposal refinement and multi-scale handling for tiny referents, rather than implying any flaw in our main filtering strategy.

\section{Self-Consistency via Multi-sample Instance Descriptions}
To reduce caption noise from the MLLM and increase robustness of the subsequent reasoning stage, we adopt a \textbf{self-consistency} procedure applied at the \emph{instance-description} level. Concretely, for each candidate region the MLLM is queried \(n=5\) times to produce five independent local descriptions. These multiple descriptions are then aggregated by the LLM: the LLM ingests the set of five descriptions for a single instance and outputs a \emph{final consolidated description} for that instance by reconciling inconsistencies and selecting the most consistent attributes across samples. The candidate selection step then proceeds using these consolidated instance descriptions.

This instance-level sampling + LLM-aggregation differs from typical output-level ensembling: we perform stochastic sampling at the multimodal-caption generation stage and rely on the LLM to perform semantic fusion, which effectively filters out hallucinations and unstable attribute mentions from individual MLLM outputs. Importantly, this does not modify any model weights and remains fully training-free.

Empirically, applying this procedure with \(n=5\) samples per instance increases RefCOCO-val accuracy from 67.1\% to 68.5\%, confirming that (1) caption noise is a dominant source of error, and (2) aggregating multiple MLLM samples with an LLM is an effective, lightweight mitigation strategy that improves grounding robustness without extra training.

\section{Candidate Selection based on LLMs}
To further analyze the impact of our candidate selection strategy, we conducted additional experiments examining different captioning strategies for guiding candidate selection, as shown in Table 2. Specifically, we tested three configurations: (1) directly using the original textual query as the caption for the target instance (Caption $\rightarrow$ Query), (2) generating captions combining the query and global image caption information (Caption $\rightarrow$ Query+Caption), and (3) using the captions generated solely from the multimodal model (Caption).
The results indicate that directly employing the original query (\textit{Caption $\rightarrow$ Query}) achieves the highest overall accuracy (around 90.6\%), significantly outperforming the scenario of using purely multimodal-generated captions (Caption). In contrast, accuracy notably decreases when solely relying on the MLLM-generated captions (average accuracy drops to 65.1\%), highlighting that inaccuracies or hallucinations introduced by the MLLM significantly impair the effectiveness of the subsequent LLM-based candidate selection step.
Moreover, integrating global captions into query-based captioning (Caption $\rightarrow$ Query+Caption) shows intermediate performance (average accuracy 85.0\%), suggesting global captions provide valuable contextual constraints yet still introduce noise due to multimodal hallucinations. These findings underscore the importance of accurate textual inputs in our reasoning framework and identify improving caption generation as a critical avenue for future research.
In summary, our ablation experiments confirm the robustness of the proposed candidate generation pipeline, clearly indicating that performance bottlenecks primarily originate from MLLM-generated captions rather than the reasoning mechanism itself.

\section{Stability Across Independent Runs}
To evaluate the robustness of our training-free pipeline, we conduct three independent runs across the RefCOCO, RefCOCO+, and RefCOCOg benchmarks. The resulting accuracy exhibits minimal variance, with an average standard deviation of approximately 0.55\% across all splits. This stability confirms that the multi-stage reasoning procedure behaves consistently under repeated executions.

\section{Segmentation Refinement Using SAM}
Beyond bounding-box grounding, we further explore the applicability of our method to referring expression segmentation by integrating a segmentation refinement module. After the LLM identifies the target bounding box, we apply a Segment-Anything-Model (SAM) to generate a segmentation mask conditioned on the selected region. This lightweight extension does not modify any part of our pipeline and requires no additional training. Empirical evaluation shows substantial improvements over prior training-free approaches: the combined system achieves 57.3 mIoU on RefCOCO-val, 51.2 mIoU on RefCOCO+-val, and 56.5 mIoU on RefCOCOg-val. These results indicate that the spatially grounded bounding boxes produced by our reasoning framework are sufficiently precise to enable effective segmentation refinement, and they highlight the versatility of the proposed architecture for downstream multimodal understanding tasks.

\section{Detailed Failure Analysis}
\begin{figure}[t]
    \centering
    \setlength{\abovecaptionskip}{0.0cm}
    \includegraphics[width=\linewidth]{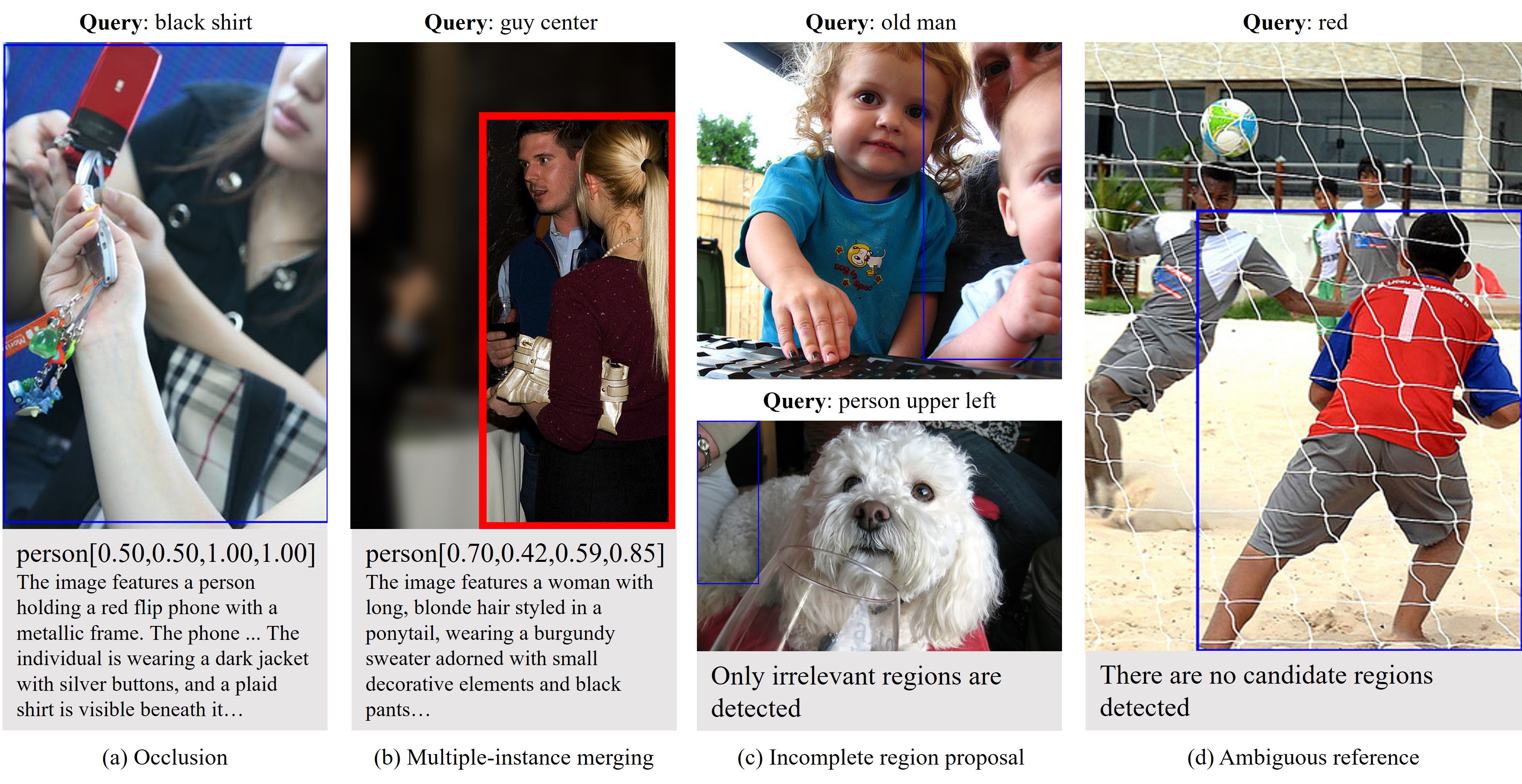}
    \caption{
    Representative failure cases. (a) \textbf{Occlusion hallucination:} confusion caused by partial occlusion. (b) \textbf{Merged instances:} separate objects grouped incorrectly. (c) \textbf{Incomplete box:} partial coverage of the target. (d) \textbf{Ambiguous query:} unclear description prevents unique identification.
    }
    \label{fig:failure_cases}
\end{figure}

\noindent
We present both qualitative and quantitative analyses of typical failure cases in Figure~\ref{fig:failure_cases}, including errors from hallucinated region descriptions and incorrect candidate rejections.

\textbf{Multimodal Generation Errors.}
We found that multimodal generation artifacts often impair caption accuracy. The MLLM sometimes produces incomplete or inaccurate descriptions due to foreground occlusion or faulty region proposals (Fig.~\ref{fig:failure_cases}a,c). In other instances, detectors merge nearby objects into a single region (Fig.~\ref{fig:failure_cases}b), confusing the captioning model. These errors can mislead the LLM and reduce grounding accuracy, indicating that future work should focus on improving the reliability and detail of multimodal captions.

\textbf{Candidate Selection Errors.}
We also observed failures in the LLM-based candidate selection step, where all options are rejected when referring expressions are ambiguous (Fig.~\ref{fig:failure_cases}d). To quantify this, we measured rejection rates on the RefCOCO+ benchmark: 0.77\% on the validation set, 0.73\% on testA, and 1.69\% on testB. These low rates confirm the robustness and effectiveness of our rejection-aware selection strategy.

\section{Implementation Details (Supplementary)}

\subsection{Coordinate Normalization}
To ensure consistent spatial representation across images of varying sizes, all bounding box coordinates are normalized. The bottom-left corner of the image is defined as the origin, and coordinates are expressed in relative terms with respect to image dimensions. Specifically, each candidate region is formatted as:
\[
(center_x, center_y, width, height)
\]
where \textit{center\_x} and \textit{center\_y} are the center coordinates of the bounding box, and \textit{width}, \textit{height} denote the box dimensions. All values are normalized to fall within $[0, 1]$ and rounded to three decimal places. This format is designed to improve parsing and interpretability by large language models (LLMs and MLLMs) during reasoning.

\subsection{Region Filtering and Selection}
To reduce noise and improve grounding quality, we filter out candidate regions with an area less than 2.5\% of the image size. This threshold is informed by an empirical analysis of the RefCOCO/+/g datasets, where no referring expression corresponds to such small regions.

After filtering detections, we apply Non-Maximum Suppression (NMS) to remove redundant boxes. Regions are then sorted by area, and the top 10 largest regions are selected as \textit{primary candidates} $\mathbf{p}_i^\text{main}$. Each primary candidate receives a detailed description $\mathbf{d}_i$ from the MLLM. The remaining candidates are only labeled with semantic concepts and relative coordinates to reduce token overhead.

As shown in Figure 3, using a confidence-based filtering approach, the average number of candidate regions larger than 2.5\% of the image area remains below 6. This supports the choice of 10 as a conservative upper bound, balancing completeness with reasoning efficiency.

\subsection{Prompts}


\begin{lstlisting}[style=promptstyle, language=Python, caption={Extracting Relevant Objects from Query.}]
VG_TEXT_GROUNDER_w_QUERY = '''\
You are a subject extractor, and you need to extract the subject from the object positioning description I give you. For example, "the painting hanging on the laptop", you need to return to me the real target subject of the sentence "painting".
Available noun examples: {noun_examples}
You need to list all possible nouns. For example, if I provide "left kid in blue shirt," your output should be "kid, child, person, shirt", with as many similar nouns as possible.
Your output should only be the extracted nouns. If there are multiple objects, you can separate them with commas. For example, "chair, person, dog".
'''
\end{lstlisting}

\begin{lstlisting}[style=promptstyle, language=Python, caption={Extracting Relevant Objects Using Query and Image Caption.}]
VG_TEXT_GROUNDER_QUERY = '''\
Image description: {image_desc}
Query: {query}
'''

VG_TEXT_GROUNDER = '''\
You are an object extractor for referred object positioning queries, and you need to extract all related objects from the object positioning query I give you.
You cannot see the image, but you can use the provided image description to more accurately extract possible involved objects when the object positioning query is very vague.

Your task: Extract all possible object names from the object positioning query given by the user. You can refer to the image description, but do not directly extract object names from the image description.

Instructions:
- Do not ask any questions to the user, even if the user's query contains questions or is ambiguous. Treat all user inputs as queries.
- You need to list all possible nouns. For example, if I provide "left kid in blue shirt," your output should be "kid, child, person, shirt", with as many similar nouns as possible.
- Your output should only be the extracted nouns. If there are multiple objects, you can separate them with commas. Do not add any extra information, such as reasoning or annotations.

Available noun examples: {noun_examples}

Example:
<user>
Image description: ... (For example, two kids playing in the park)
Query: left kid in blue shirt
<assistant>
kid, child, person, shirt, clothing
'''
\end{lstlisting}

\begin{lstlisting}[style=promptstyle, language=Python, caption={Generating Local Object Descriptions with MLLM.}]
MLLM_GLOBAL_DESC_PROMPT = "Describe the image in a few sentences."
\end{lstlisting}

\begin{lstlisting}[style=promptstyle, language=Python, caption={Generating Local Object Descriptions with MLLM.}]
MLLM_INSTANCE_DESC_PROMPT = "Describe the object marked by {visual_prompt}: {name}[{a1:.3f}, {a2:.3f}, {a3:.3f}, {a4:.3f}]. Note: The coordinates [*, *, *, *] are in the {box_format} format."

MLLM_INSTANCE_DESC_SYSTEM_PROMPT = "You are a fine-grained description multimodal model. Your task is to provide a detailed description (or captioning) of the object marked by {visual_prompt}."
\end{lstlisting}

\begin{lstlisting}[style=promptstyle, language=Python, caption={Selecting Target Bounding Box with LLM.}]
MAIN_INSTANCE_TEXT = '{idx}. {category_name}[{center_x:.3f}, {center_y:.3f}, {width:.3f}, {height:.3f}]\nInstance Description: {description}'

OTHER_INSTANCE_TEXT = '{idx}. {category_name}[{center_x:.3f}, {center_y:.3f}, {width:.3f}, {height:.3f}]'

LLM_INSTANCE_DESC = """\
## Main Instance Descriptions
{main_instance_descs}

## Other Instance Descriptions
{other_instance_descs}

IMPORTANT NOTE: The coordinates [*, *, *, *] are in the {box_format} format. The {box_format} values are normalized to the range [0, 1]. A higher y value indicates a higher/top position, while a lower value indicates a lower/bottom position. Similarly, a higher x value indicates a position to the right, and a lower value to the left. By carefully observing the center_x and center_y of an instance, you can determine its position on the image, while width and height indicate its size or proximity.
"""

LLM_SYSTEM_PROMPT_WITH_COT = """\
You are known as the "Blind Teacher," a highly intelligent educator specializing in reasoning and critical thinking.
Although you cannot see the image, you can understand it through textual descriptions. You will be provided with a description of the image and the main objects in it, and you need to answer questions based on these descriptions.
Your task: Identify the object referred to by the user from the given list of objects.

Here is the textual description:
# Image Description
{global_desc}
# Instance Description
{instance_desc}

Think step by step to identify the object referred to by the user. Carefully analyze the descriptions provided and match them with the user's query.

Your output format: Directly output the index of the object. Do not include any additional information in Answer.

Example:

Input:
User: The person wearing red clothes
Output:
Reasoning Step 1: ...(Identify the person and red clothes in the image. Analyze the descriptions provided.)
Reasoning Step 2: ...(Identify the person wearing red clothes. Analyze the descriptions provided.)
...
Answer: 1
"""
\end{lstlisting}

\section{More Examples}

\begin{figure*}[ht]
\centering
\includegraphics[width=\linewidth]{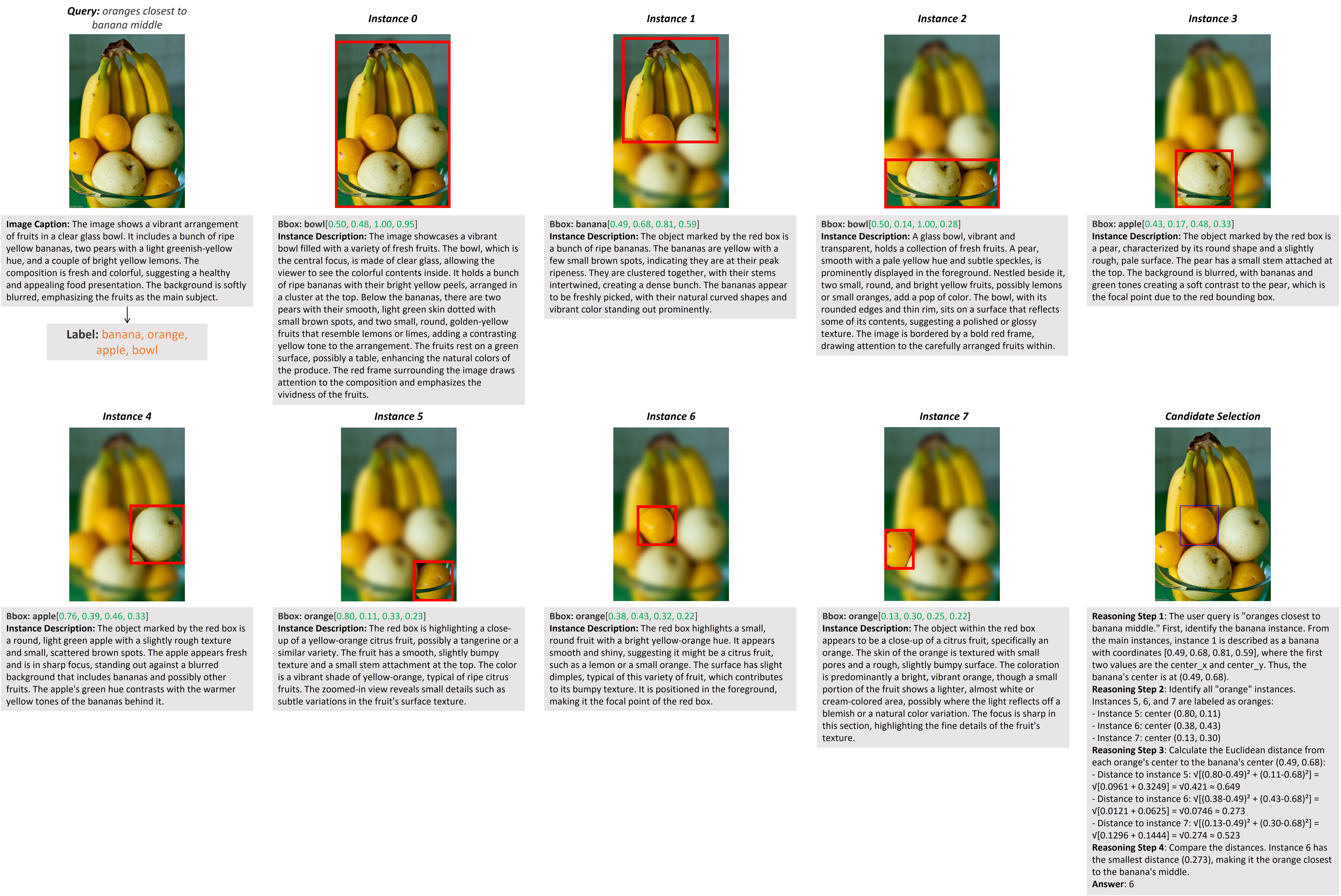}
\caption{Qualitative example illustrating how our \textit{GroundingAgent} method effectively handles spatial relationships and fine-grained visual descriptions. Given the query \textit{``oranges closest to banana middle''}, our framework first generates multiple candidate bounding boxes for oranges and bananas, each enriched with detailed semantic descriptions provided by the MLLM. Subsequently, the LLM systematically analyzes spatial coordinates and descriptive semantics, explicitly reasoning through the proximity of each orange to the middle banana to identify the closest instance correctly. } 
\label{fig:case2}
\end{figure*}

\begin{figure*}[h]
\centering
\includegraphics[width=\linewidth]{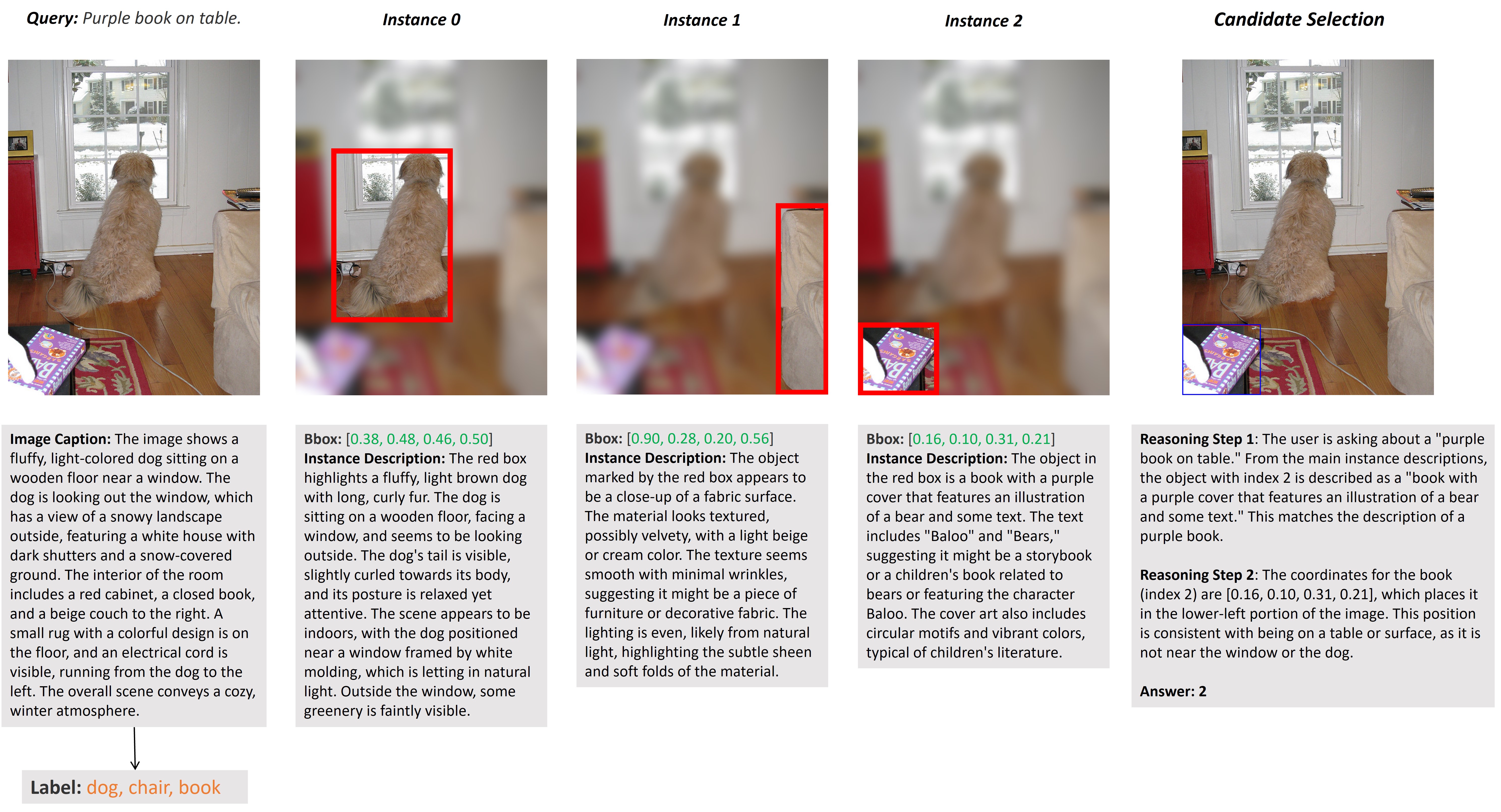}
\caption{Example of fine-grained object description}
\label{fig:example1}
\end{figure*}


\begin{figure*}[h]
\centering
\includegraphics[width=\linewidth]{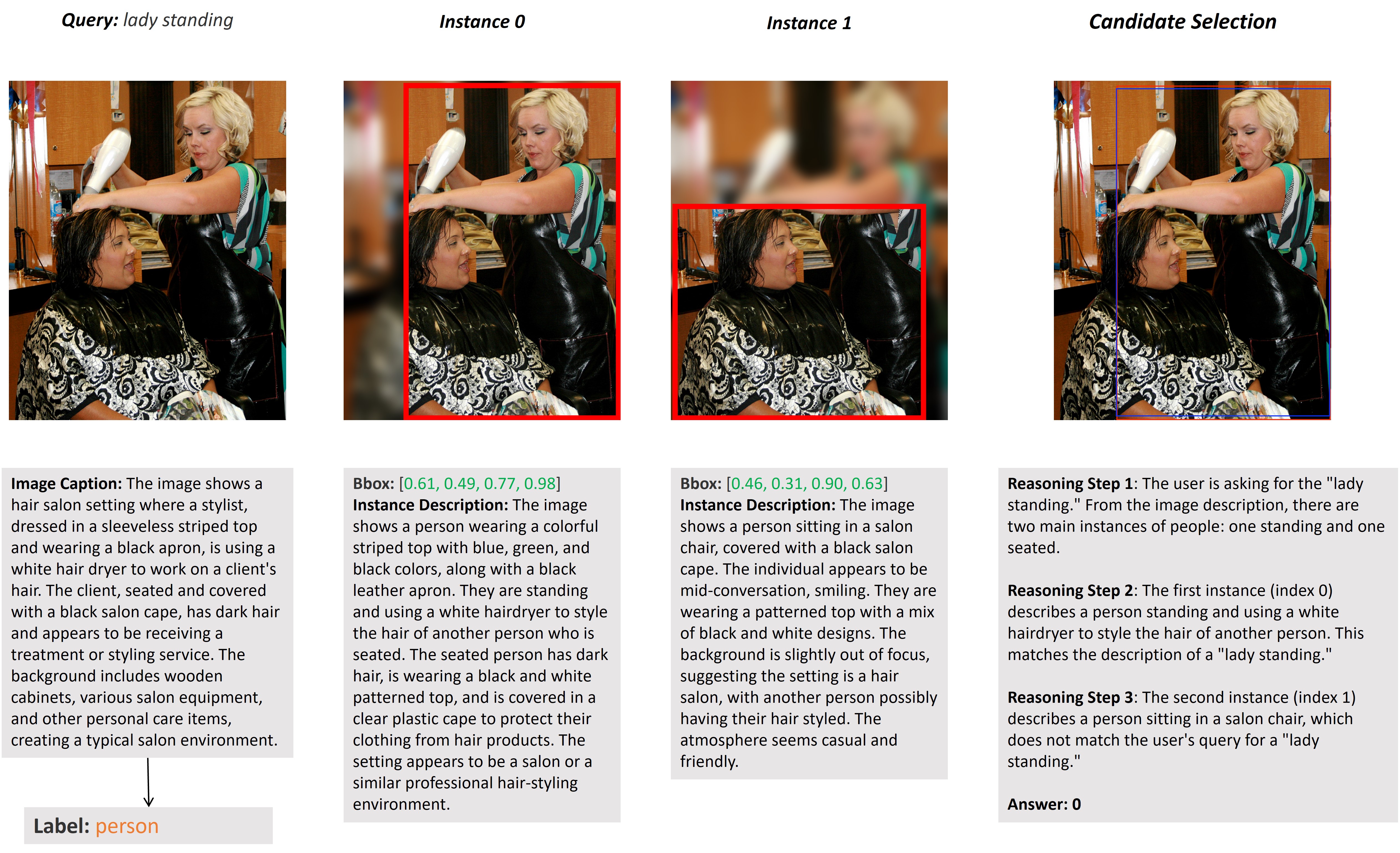}
\caption{Example of query involving human actions}
\label{fig:example3}
\end{figure*}

\begin{figure*}[h]
\centering
\includegraphics[width=\linewidth]{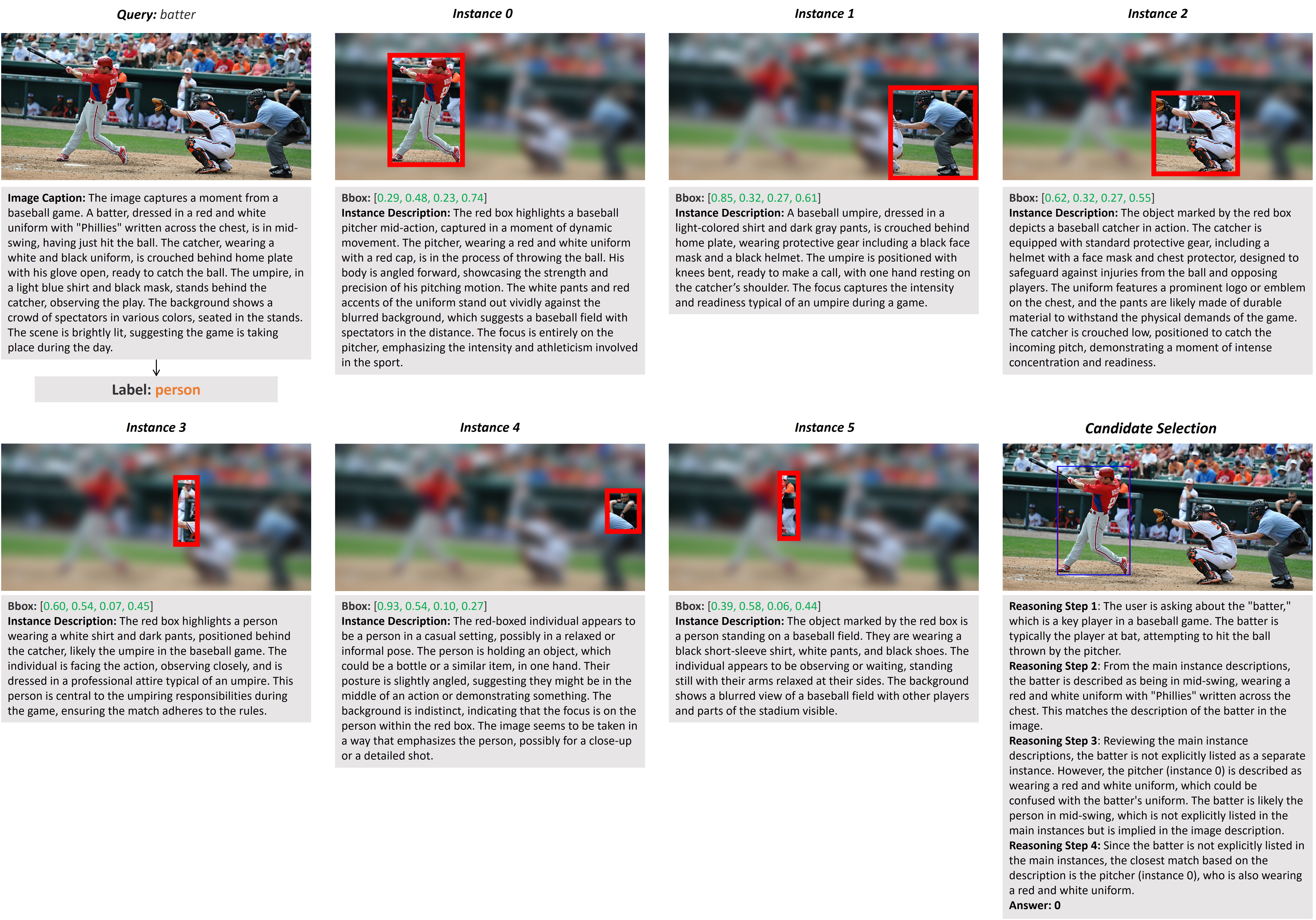}
\caption{LLM's reasoning capability correcting MLLM's description errors}
\label{fig:example4}
\end{figure*}

\begin{figure*}[h]
\centering
\includegraphics[width=\linewidth]{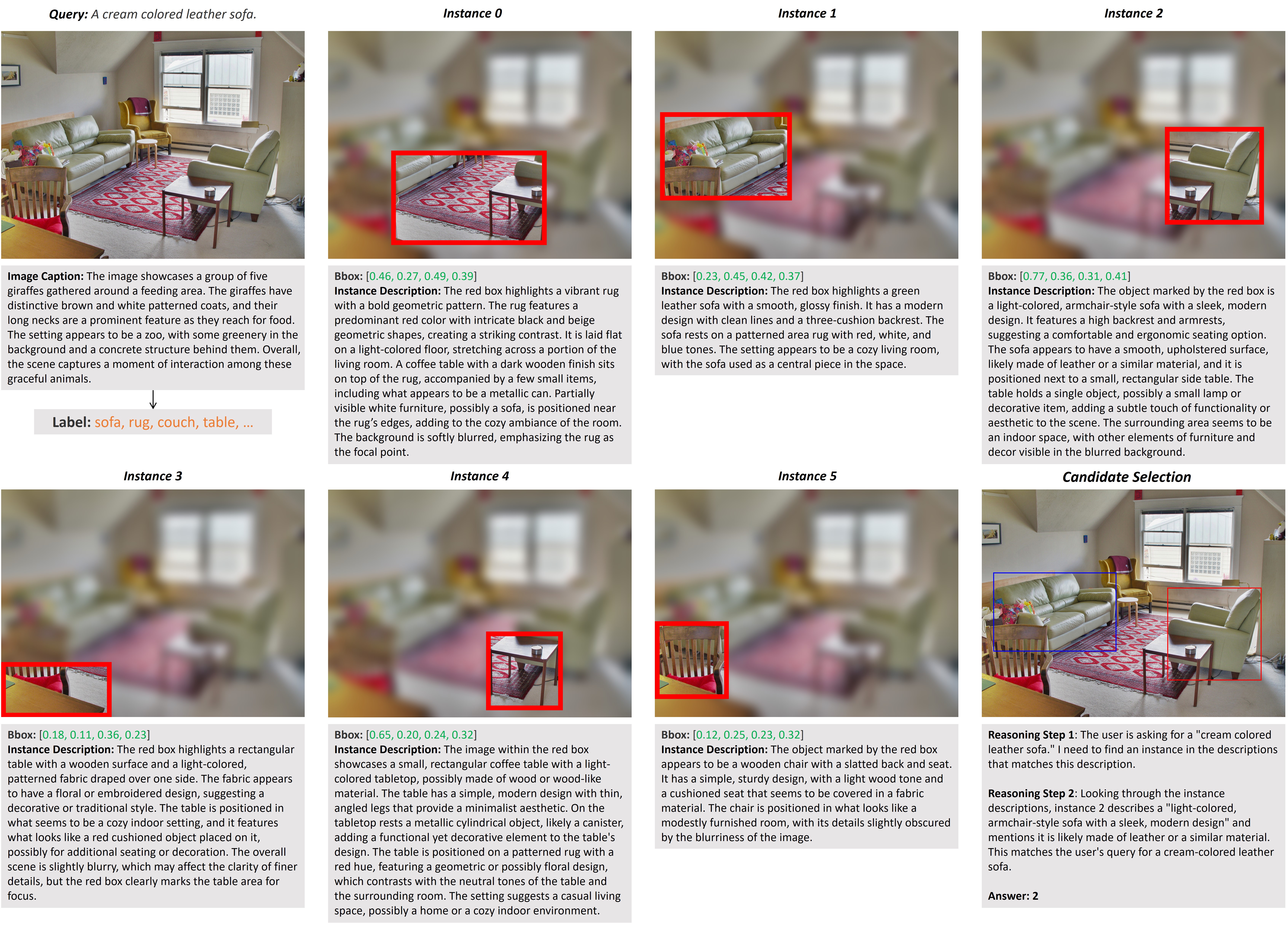}
\caption{An example of failure case}
\label{fig:example5}
\end{figure*}

\end{document}